\documentclass[letterpaper, 10 pt, conference]{ieeeconf}  

\IEEEoverridecommandlockouts                              

\usepackage{graphicx}
\usepackage{times} 
\usepackage{amsmath} 
\usepackage{amssymb}  
\usepackage{soul}
\usepackage{xcolor}
\usepackage{cite}

\usepackage{mathcomp}

\title{\LARGE \bf
Transporting Unsecured Stacked Payloads with a Quadrupedal Robot via Multi-Objective Reinforcement Learning
}

\author{Nobuo Namura$^{1,*}$, Masayuki Hiromoto$^{1,*}$, Kento Uemura$^{1}$,
Hironobu Sasaki$^{1}$, and Kanata Suzuki$^{1}$%
\thanks{$^{*}$These authors contributed equally.}%
\thanks{$^{1}$All authors are with Fujitsu Limited, Kawasaki, Japan.
Nobuo Namura is the corresponding author.
E-mail: {\tt\small namura.nobuo@fujitsu.com}}%
}

\begin{document}

\maketitle
\thispagestyle{empty}
\pagestyle{empty}

\begin{abstract}
Transporting unsecured payloads with legged robots over uneven terrain requires balancing locomotion performance and payload stability, since aggressive motion can destabilize the payload even when the robot remains stable.
We study quadrupedal transportation of unsecured stacked boxes on an edgeless torso-mounted board without dedicated payload sensors or active carrier mechanisms.
To address this trade-off, we propose Payload-Adaptive Multi-Objective Reinforcement learning for Transportation (PAMORT).
PAMORT trains a multi-objective base policy conditioned on a preference vector that weights locomotion and payload-stability reward groups, then trains a weight adjuster on the frozen policy to adapt this preference online from proprioception.
In simulation, PAMORT achieves comparable or better overall transportation success than a corresponding single-objective baseline across different payload configurations, including an unseen three-box stack, despite training only with two boxes.
Real-world experiments on a Unitree Go2 demonstrate zero-shot transfer to slopes and steps at or beyond the training difficulty, with mean success rates of 0.850 for PAMORT and 0.675 for the baseline across eight tasks.
These results demonstrate robust unsecured-payload transportation with online adaptation of the locomotion--payload trade-off from proprioceptive information.
\end{abstract}

\section{INTRODUCTION}
\label{sec:introduction}
Recent advances in reinforcement learning have enabled quadrupedal robots to traverse challenging terrain, including stairs, slopes, and rough ground. As locomotion becomes more robust, practical tasks beyond mobility itself are becoming increasingly important. One such task is payload transportation, for which legged robots are particularly attractive in environments where wheeled platforms cannot easily operate.

\begin{figure}[t]
    \centering
    \includegraphics[width=\linewidth]{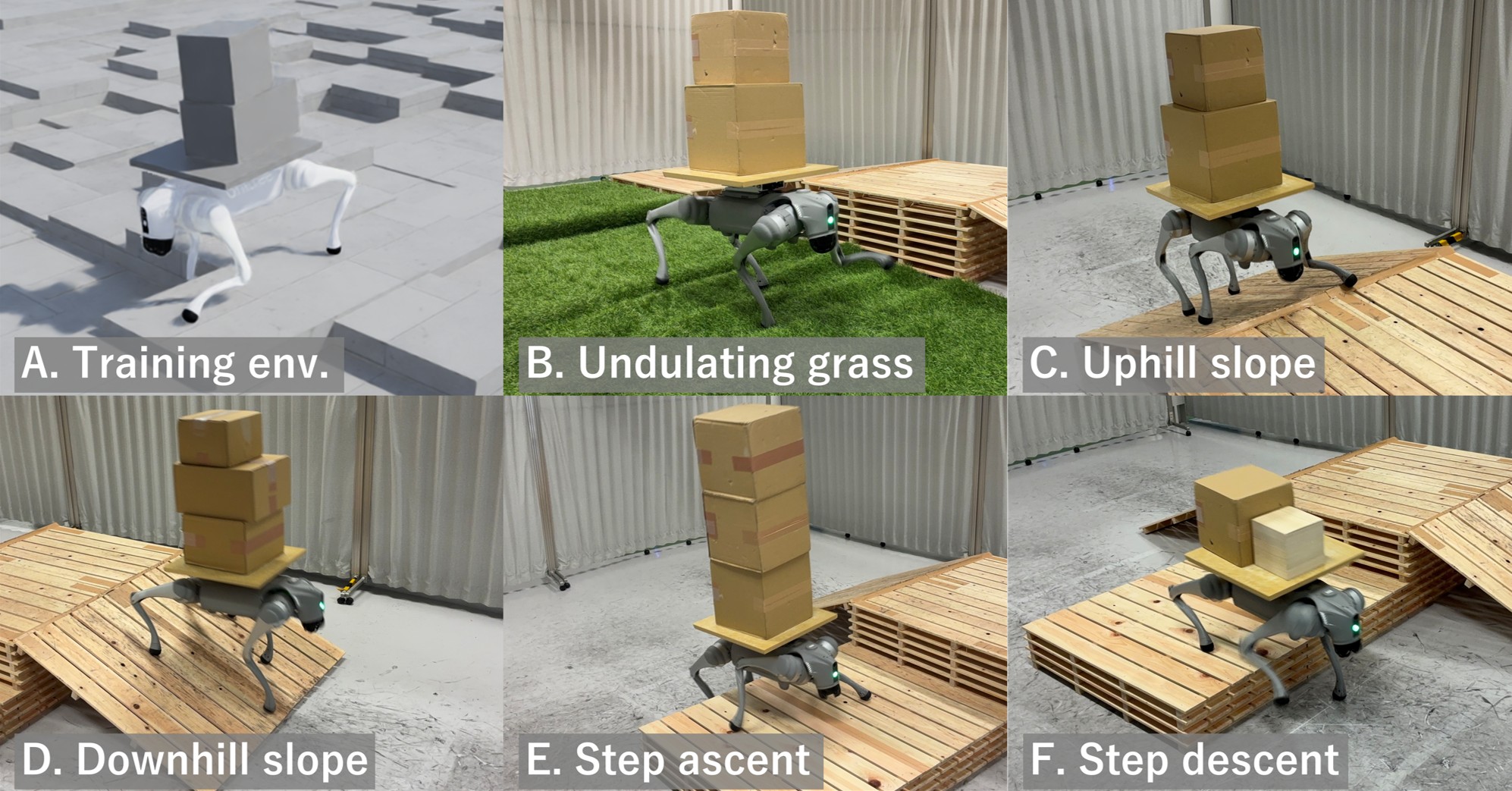}
    \caption{
    Unsecured payload transportation over uneven terrain. The training task uses two unsecured stacked boxes on an edgeless carrier board. \textbf{(A)} Simulated training environment. \textbf{(B, C)} Real-world examples with the training payload configuration. \textbf{(D--F)} Real-world examples with payload configurations not used during training: a three-box stack in \textbf{(D, E)} and two side-by-side boxes in \textbf{(F)}.
    }
    \label{fig:task}
\end{figure}

Conventional robotic transportation often mechanically secures the payload to the robot or constrains it using sidewalls, containers, or other fixtures. Such approaches are effective for reliable transport, but they may require payload-specific fastening, loading, and unloading procedures or dedicated carrier structures. In applications such as warehouses or construction sites, however, it can be useful to temporarily transport boxes, tools, or materials without securing each item individually. Rather than replacing mechanically secured transportation, this work aims to complement it with a more flexible capability that does not rely on payload-specific fixation. We therefore consider the task shown in Fig.~\ref{fig:task}, where a quadruped carries unsecured stacked boxes on a flat carrier while traversing uneven terrain.

Transporting unsecured payloads introduces challenges beyond locomotion with rigidly attached loads. Even when the robot itself remains stable, acceleration, body motion, and terrain-induced impacts can cause the payloads to slide, become misaligned, or fall. Payload stabilization therefore requires state-dependent whole-body motions and cannot, in general, be reduced to simply keeping the torso level. Designing such coupled behaviors manually is difficult, making reinforcement learning well suited to this problem. At the same time, the task exhibits an inherent trade-off between locomotion performance and payload stability. Aggressive velocity tracking can increase payload motion, whereas prioritizing payload stability can lead to overly conservative locomotion and reduced traversal performance.

To address this trade-off, we propose Payload-Adaptive Multi-Objective Reinforcement learning for Transportation (PAMORT), a framework that adapts the balance between locomotion performance and payload stability online. PAMORT builds on TAR~\cite{mousa2025tar}, a learning-based locomotion framework using privileged teacher representations and a proprioceptive student encoder for terrain adaptation. We extend TAR to a multi-objective (MO) base policy that separates the reward into \emph{move} and \emph{payload} groups and conditions a single policy on their objective weights, enabling behaviors from conservative payload preservation to aggressive locomotion. A weight adjuster is then trained on the frozen MO base policy to adapt these weights online from the robot state. PAMORT therefore adjusts the locomotion--payload trade-off without dedicated payload sensors, manipulators, or active carrier mechanisms. We also introduce a preference-aware terrain curriculum to facilitate learning across different objective balances.

In simulation, the MO base policy represents different \emph{move}--\emph{payload} trade-offs within a single policy, and PAMORT achieves higher overall transportation success than the task-adapted TAR baseline (TAR*) across diverse uneven terrains. Its advantage is also retained when the number of payloads differs from the two-payload training configuration. Real-world experiments on a Unitree Go2 further demonstrate that unsecured stacked payloads can be transported over uneven terrain, while preserving the performance advantage of PAMORT observed in simulation.

The main contributions of this work are:
\begin{itemize}
\item We study uneven-terrain transportation of unsecured stacked payloads and demonstrate that the task can be performed using only the quadruped's standard proprioceptive sensing and leg degrees of freedom, without dedicated payload sensors, manipulators, or active carrier mechanisms.
\item We propose PAMORT, a payload-adaptive multi-objective reinforcement learning framework that combines a preference-conditioned MO base policy with an online weight adjuster to explicitly manage the trade-off between locomotion performance and payload stability, supported by a preference-aware terrain curriculum for effective training across the preference space.
\item Through simulation and real-world experiments, we show that PAMORT represents multiple locomotion--payload trade-offs within a single policy and improves transportation performance over the TAR* baseline while generalizing to payload configurations different from those used during training.
\end{itemize}

\section{RELATED WORK}

\subsection{Payload Transportation with Legged Robots}
Many studies on payload transportation with legged robots assume that the payload is rigidly attached to the robot or mechanically constrained using carriers, containers, or high-friction interfaces~\cite{jin2022unknown,youngquist2025rina,arnold2025leva,chang2025beyond,kurva2025mule}. In these settings, the main challenge is typically to maintain robust locomotion under changes in total mass, center of mass, or inertial properties. Mechanical constraint is effective for reliable transportation, although it may require payload-specific fastening, loading procedures, or dedicated carrier structures.

Other works have considered unsecured or dynamically moving payloads. Ball-on-plate stabilization has been studied with legged robots~\cite{chen2019ball}, while quadruped-mounted manipulators have been used for active payload stabilization~\cite{morlando2022nonprehensile}. LocoTouch~\cite{lin2025locotouch} uses tactile sensing to directly observe payload--carrier interaction during locomotion. Dynamic loads moving inside containers have also been investigated~\cite{chang2025beyond,kurva2025mule}, and SteadyTray considers stabilization of objects carried by a humanoid robot~\cite{huang2026steadytray}. These studies show that unstable payloads can be handled through mechanical constraint, direct sensing, or additional actuation.

Compared with a single unsecured object, stacked payloads introduce not only motion relative to the carrier but also inter-payload displacement and toppling, making them a compact but challenging case of unsecured transportation. However, uneven-terrain transportation of such payloads by a quadruped without dedicated payload sensing or additional stabilization hardware remains underexplored. Our work focuses on this setting using only standard proprioception and leg actuation.

\subsection{Multi-Objective Reinforcement Learning}
Conventional reinforcement learning typically combines multiple criteria into a single scalar reward using fixed weights and optimizes a policy for one predefined objective balance. Multi-objective reinforcement learning (MORL), in contrast, explicitly represents multiple objectives and their trade-offs, and has been studied through approaches such as learning sets of Pareto-optimal policies~\cite{roijers2013survey,xu2020prediction}.

Preference-conditioned MORL instead incorporates an objective-weight or preference vector into a single policy, allowing different trade-offs to be selected after training by varying the preference input~\cite{abels2019dynamic,yang2019generalized,basaklar2023pd}. This formulation is particularly attractive for robot control, where priorities such as speed, stability, safety, and energy efficiency may conflict and change with the operating condition. Related conditioning schemes have also been explored in continuous control and locomotion to generate diverse behaviors from a single policy~\cite{margolis2023walk,leng2025preference}.

More recent work has further considered adapting preferences online rather than selecting them manually. AMOR~\cite{alegre2025amor}, for example, learns a high-level policy that dynamically adjusts the objective weights of a frozen preference-conditioned controller for character motion tracking, enabling the behavior to adapt to changing task requirements.

Our work follows this direction but considers unsecured payload transportation, where locomotion performance and payload stability must be balanced under partial observability. The objective weights are adapted online without direct observation of the payload state.

\section{METHOD}
\label{sec:method}

\begin{figure}[t]
    \centering
    \includegraphics[width=\linewidth]{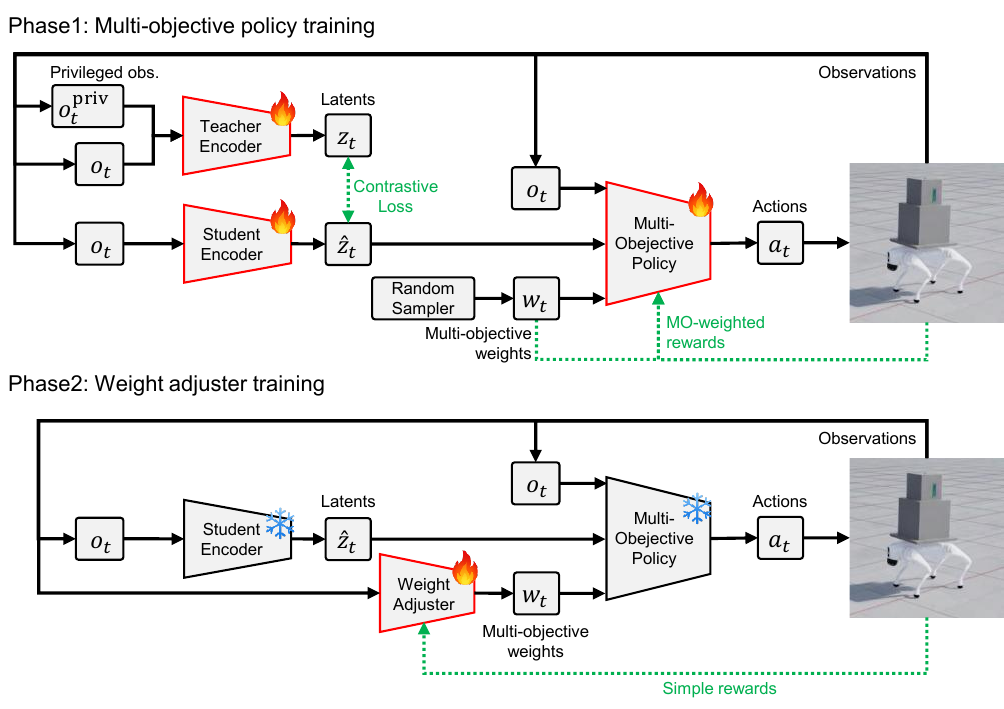}
    \caption{Overview of PAMORT's two-stage training framework. In Phase~1, a preference-conditioned MO policy is trained with grouped rewards and sampled preferences. In Phase~2, the MO policy is frozen and the weight adjuster is trained to adapt the preference online. TAR~\cite{mousa2025tar}'s representation and velocity-estimation details are omitted for clarity.}
    \label{fig:overview}
\end{figure}

Our locomotion backbone follows proximal policy optimization (PPO)~\cite{schulman2017proximal} with asymmetric actor--critic (AAC) training~\cite{pinto2018asymmetric}, where privileged information is available only during training.
Among latent-adaptation approaches such as RMA~\cite{kumar2021rma} and concurrent teacher--student learning in CTS~\cite{wang2024cts}, we adopt TAR~\cite{mousa2025tar} because its contrastive alignment provides a suitable mechanism for implicitly encoding privileged payload-related state information into a proprioceptive latent representation.
During training, TAR's teacher encoder and the AAC critic share the same privileged observation, while the student encoder uses only proprioceptive information.
We extend TAR to the multi-objective stacked-payload transportation task considered here. Let $o_t$ denote the policy observation at time $t$; the policy uses only proprioceptive signals at deployment, without direct payload-state observations.
As shown in Fig.~\ref{fig:overview}, PAMORT first trains a preference-conditioned MO base policy and then trains a weight adjuster on the frozen policy.

\subsection{Phase~1: Multi-Objective Policy Training}

In the first phase, TAR~\cite{mousa2025tar} is used as the backbone of the PPO-based actor--critic framework. The TAR components, including the latent representation and the velocity estimation branch, are kept intact, together with the triplet loss and the velocity loss. This framework is extended by conditioning the actor on an MO weight vector \(\boldsymbol{w}_t\), which makes the policy explicitly preference-conditioned.
The policy is written as
\begin{equation}
a_t \sim \pi_\theta\!\left(a_t \mid o_t, \hat{z}_t, \hat{v}_t, \boldsymbol{w}_t\right),
\label{eq:policy_phase1}
\end{equation}
where \(\hat{z}_t\) is the student latent representation produced by TAR, \(\hat{v}_t\) is the estimated velocity, and \(\boldsymbol{w}_t\) is the MO weight vector. The critic uses a multi-head structure and predicts one value head per objective group.

\subsubsection{Reward Grouping}
The scalar reward terms are divided into two objective groups,
\texttt{move} and \texttt{payload}, corresponding to locomotion performance
and payload preservation, respectively.
Their summed rewards form the reward vector
$\boldsymbol{r}_t=[r_t^{\mathrm{move}},r_t^{\mathrm{payload}}]^\top$.
The assignment of individual reward terms is given in Table~\ref{tab:reward}.

\subsubsection{Preference Sampling}
The policy is conditioned on a preference vector
\begin{equation}
\boldsymbol{w}=[w^{\mathrm{move}},w^{\mathrm{payload}}]^\top\in\Delta^1,
\end{equation}
where $\Delta^1:=\{\boldsymbol{w}\in\mathbb{R}_{\geq0}^2 \mid \mathbf{1}^\top\boldsymbol{w}=1\}$ is the one-dimensional simplex.
The preference specifies the trade-off between the two objectives.
During training, $\boldsymbol{w}$ is sampled uniformly over the preference simplex as
$\boldsymbol{w}\sim\mathrm{Dirichlet}(1,1)$
and remains fixed throughout each episode.

\subsubsection{Multi-Objective PPO}
PAMORT retains the original TAR teacher--student representation-learning objectives and extends its policy optimization to multiple objectives.
For each reward group, PAMORT maintains an objective-specific value head and computes the corresponding advantage independently.
Before return and advantage computation, each reward group is standardized independently using its own running statistics.
The resulting advantage vector
$\boldsymbol{A}_t=[A_t^{\mathrm{move}},A_t^{\mathrm{payload}}]^\top$
is aggregated according to the episode-wise preference as
$\tilde{A}_t = \boldsymbol{w}^{\top}\boldsymbol{A}_t$.
The scalar $\tilde{A}_t$ is used in the standard PPO policy objective~\cite{schulman2017proximal}, while each value head is optimized using its corresponding return.

\subsubsection{Preference-Aware Curriculum}
Terrain difficulty is controlled by a curriculum during training.
A conventional distance-based curriculum is unsuitable for preference-conditioned training because low-\texttt{move} preferences intentionally induce more conservative behaviors with shorter travel distances.
Such behavior can be misinterpreted as insufficient training progress, causing the terrain level to decrease even when the policy is behaving consistently with the sampled preference.
We therefore define objective-specific competences and combine them into an aggregated competence score using a preference-weighted geometric mean:
\begin{align}
c_{\mathrm{agg}}
&=
\exp\Big(
w_t^{\mathrm{move}} \log \max(c_{\mathrm{move}}, c_{\min})
\nonumber\\
&\qquad\qquad
+
w_t^{\mathrm{payload}} \log \max(c_{\mathrm{payload}}, c_{\min})
\Big),
\label{eq:pref_competence}
\end{align}
where $c_{\mathrm{move}}\in[0,1]$ measures normalized locomotion progress and $c_{\mathrm{payload}}\in[0,1]$ measures the progress reached before the first payload drop, or the final progress if no drop occurs; we use $c_{\min}=0.05$. Terrain difficulty is increased when $c_{\mathrm{agg}}\geq0.8$ and decreased when $c_{\mathrm{agg}}\leq0.5$, with a robot fall also triggering a level decrease.

\subsection{Phase~2: Weight Adjuster Training}
\label{sec:wa}

In the second training phase, the pretrained MO base policy is frozen and WA is trained to adapt the preference online according to the current state.
At each WA update, it outputs logits
\begin{equation}
\boldsymbol{u}_t = f_\phi\!\left(o_t,\ \hat{z}_t,\ \hat{v}_t,\ \boldsymbol{w}_{t-1}\right),
\label{eq:wa_output}
\end{equation}
which are converted to a valid preference by
$\boldsymbol{w}_t=\mathrm{softmax}(\boldsymbol{u}_t)$
and used to condition the frozen MO base policy.

To avoid unstable switching between objective balances, WA updates the preference at a lower frequency than the low-level policy and holds it fixed between updates.
A smoothness loss further penalizes abrupt changes between consecutive preferences:
\begin{equation}
    L_{\mathrm{smooth}}
    =
    \lambda_t
    \left\|
        \boldsymbol{w}_t-\boldsymbol{w}_{t-1}
    \right|_2^2,
    \quad
    \lambda_t\in[0.02,0.5].
    \label{eq:smoothness}
\end{equation}
The coefficient $\lambda_t$ is increased in stable states and reduced during rapid command or posture changes, encouraging gradual preference adaptation while allowing WA to respond quickly when the operating condition changes.

\subsection{Reward Design} 
\label{sec:reward}
In addition to standard uneven-terrain locomotion rewards, we introduce four task-specific terms for stacked-payload stability. We use different reward configurations for TAR*, PAMORT's MO base policy, and WA according to their training objectives.

\begin{table}[t]
\caption{Reward configurations}
\label{tab:reward}
\centering
\setlength{\tabcolsep}{5pt}
\begin{tabular}{l c c c c}
\hline
& & & \multicolumn{2}{c}{PAMORT} \\
\cline{4-5}
Reward term & Weight & TAR* & MO group & WA \\
\hline
Lin. velocity tracking
& $1.5$ & \checkmark & \texttt{move} & \checkmark \\

Ang. velocity tracking
& $0.75$ & \checkmark & \texttt{move} & \checkmark \\

Lin. velocity ($z$)
& $-2.0$ & \checkmark & \texttt{move} & -- \\

Ang. velocity ($xy$)
& $-0.05$ & \checkmark & \texttt{move} & -- \\

Joint velocity
& $-1.0{\times}10^{-3}$ & \checkmark & \texttt{move} & -- \\

Joint acceleration
& $-2.5{\times}10^{-7}$ & \checkmark & \texttt{move} & -- \\

Joint torque
& $-2.0{\times}10^{-4}$ & \checkmark & \texttt{move} & -- \\

Action rate
& $-0.1$ & \checkmark & \texttt{move} & -- \\

Joint limit violation
& $-10.0$ & \checkmark & \texttt{move} & -- \\

Soft-limit ratio
& $-1.0$ & \checkmark & \texttt{move} & -- \\

Joint position
& $-0.7$ & \checkmark & \texttt{move} & -- \\

Joint energy
& $-2.0{\times}10^{-5}$ & \checkmark & \texttt{move} & -- \\

Feet air time
& $0.1$ & \checkmark & \texttt{move} & -- \\

Air-time variance
& $-1.0$ & \checkmark & \texttt{move} & -- \\

Feet slide
& $-0.1$ & \checkmark & \texttt{move} & -- \\

Feet contact forces
& $-0.02$ & \checkmark & \texttt{move} & -- \\

Undesired contacts
& $-1.0$ & \checkmark & \texttt{move} & -- \\

Bottom-box alignment
& $2.0$ & \checkmark & \texttt{payload} & \checkmark \\

Top-box alignment
& $2.0$ & \checkmark & \texttt{payload} & \checkmark \\

Bottom-box planar accel.
& $-2.5$ & \checkmark & \texttt{payload} & \checkmark \\

Top-box planar accel.
& $-0.25$ & \checkmark & \texttt{payload} & \checkmark \\

Bottom-box drop
& $-150$ & -- & \texttt{payload} & -- \\

Top-box drop
& $-100$ & -- & \texttt{payload} & -- \\
\hline
\end{tabular}
\end{table}

\subsubsection{Alignment Rewards}
To suppress payload displacement and relative misalignment, we introduce alignment rewards based on the displacement of each box from its support. 
We define $\boldsymbol{d}^{\mathrm{bottom}}_{xy}$ as the planar offset of the bottom box from the board center and $\boldsymbol{d}^{\mathrm{top}}_{xy}$ as the planar offset of the top box from the bottom-box center. 
The alignment rewards are
\begin{equation}
r_\mathrm{align}^i
=
1-
\mathrm{clip}\!\left(
\frac{\lVert \boldsymbol{d}^{i}_{xy}\rVert_2}{d_{\max}},
0,1
\right),
\quad
i\in\{\mathrm{bottom},\mathrm{top}\},
\label{eq:alignment_rewards}
\end{equation}
where
$d_{\max}=\sqrt{(L_x^{\mathrm{board}}/2)^2+(L_y^{\mathrm{board}}/2)^2}$
is the planar distance from the board center to a corner.

\subsubsection{Planar Acceleration Penalties}

Alignment rewards measure payload displacement but do not distinguish motion away from or back toward the support center.
We therefore penalize planar acceleration while allowing restoring motion toward the center.

For each box $i\in\{\mathrm{bottom},\mathrm{top}\}$, let $\boldsymbol{a}_{xy}^{i}$ and $\boldsymbol{g}_{xy}^{i}$ denote its planar acceleration and the planar component of gravity, respectively, in the reference frame of its supporting body. 
We define the normalized residual planar acceleration as
$\boldsymbol{u}^{i}
=
({\boldsymbol{g}_{xy}^{i}-\boldsymbol{a}_{xy}^{i}})/{\lVert\boldsymbol{g}\rVert_2}$,
and the target acceleration for restoring payload alignment as
$\boldsymbol{u}_{\star}^{i}
=
-(k/\lVert\boldsymbol{g}\rVert_2)(\boldsymbol{d}_{xy}^{i}/d_{\max})$,
where $\boldsymbol{g}$ denotes the gravitational acceleration vector and $k=2\,\mathrm{m/s^2}$ sets the target restoring acceleration magnitude at $d_{\max}$. Minimizing $\lVert\boldsymbol{u}^{i}\rVert_2$ encourages the robot to adjust its posture to counteract payload acceleration caused by locomotion, including acceleration, deceleration, and turning. The target $\boldsymbol{u}_{\star}^{i}$ points toward the support center and encourages torso tilt that restores payload alignment.
Using these quantities, the planar acceleration penalty is defined as
\begin{equation}
p_{\mathrm{acc}}^i
=
\left[
\frac{1}{2}
\left(
\lVert \boldsymbol{u}^i\rVert_2
+
\lVert \boldsymbol{u}^i-\boldsymbol{u}_{\star}^i\rVert_2
\right)
\right]^2.
\label{eq:planar_acc}
\end{equation}
This formulation balances suppression of unnecessary planar motion with restoring motion for displaced payloads.

\subsubsection{Reward Configurations}
The original TAR~\cite{mousa2025tar} is designed for locomotion without payloads. For comparison with PAMORT, we therefore define a task-adapted TAR baseline, denoted TAR*, by augmenting TAR with the stacked-payload-specific reward terms and combining all terms into a single scalar reward with fixed weights. In TAR*, an episode terminates when either box is dropped, so no explicit box-drop penalty is used.

PAMORT uses the same reward terms but assigns them to the \texttt{move} and \texttt{payload} groups.
During MO base policy training, payload drops do not terminate the episode. Instead, one-time penalties of $-150$ and $-100$ are applied for the first drop of the bottom and top boxes, respectively, and the corresponding dense payload terms are masked after a drop. This allows move-dominant preferences to continue learning locomotion after losing the payload.

For WA training, the pretrained MO base policy is frozen and only the preference selection is optimized.
WA therefore uses a lighter scalar reward containing velocity tracking and the stacked-payload-specific terms.
Unlike the MO base policy, a payload drop terminates the WA-training episode, as in TAR*.

Table~\ref{tab:reward} summarizes all the reward configurations.

\section{TRAINING SETUP}
\subsection{Simulation Environment}

We train all policies with parallel Unitree Go2 environments in Isaac Lab.
Each robot has a rigid, edgeless carrier board with
$L_x^{\mathrm{board}}\times L_y^{\mathrm{board}}=0.4\,\mathrm{m}\times0.4\,\mathrm{m}$
and thickness $0.02\,\mathrm{m}$ fixed to its torso.
TAR* and PAMORT's WA use 8,192 environments on four GPUs, while PAMORT's MO base policy uses twice the resources with 16,384 environments on eight GPUs to cover the additional preference dimension within a single policy.
The simulation time step is $5\,\mathrm{ms}$ with a control decimation of 4, yielding a $50\,\mathrm{Hz}$ control frequency and a maximum episode length of $20\,\mathrm{s}$.
Velocity commands are uniformly sampled as $v_x\in[-1.0,1.0]\,\mathrm{m/s}$, $v_y\in[-0.5,0.5]\,\mathrm{m/s}$, and $\omega_z\in[-1.0,1.0]\,\mathrm{rad/s}$.

We use procedurally generated $8\,\mathrm{m}\times8\,\mathrm{m}$ terrain patches with 10 difficulty levels, including flat terrain and the same uneven-terrain families used for evaluation (Sec.~\ref{sec:sim} and Table~\ref{tab:eval0_terrains}).
Terrain curricula are used for TAR* and PAMORT's MO base policy, with PAMORT following the preference-aware curriculum in Sec.~\ref{sec:method}\@.
WA training uses no terrain curriculum.

Domain randomization settings are summarized in Table~\ref{tab:dr_settings}\@.
We randomize robot and payload properties during training for sim-to-real robustness and additionally apply intermittent planar velocity perturbations to the robot and both payload boxes.

\begin{table}[t]
  \centering
  \caption{Domain randomization settings}
  \label{tab:dr_settings}
  \begin{tabular}{lll}
    \hline
    Parameter & Range & Unit \\
    \hline
    Foot friction (static, dynamic) & $[0.3,\,1.2]^2$ & -- \\
    Foot restitution & $[0.0,\,0.15]$ & -- \\
    Base mass offset & $[-1.0,\,3.0]$ & kg \\
    Actuator \& PD gain factors & $[0.9,\,1.1]^{12 \times 3}$ & -- \\
    Action delay & $[0.0,\,0.01]$ & s \\
    Bottom box size $(x,y,z)$ & $[15,39]^2\times[7.5,45]$ & cm \\
    Top box size $(x,y,z)$ & $[10,30]^2\times[5,30]$ & cm \\
    Box mass (bottom, top) & $[0.2,4.0],\ [0.1,2.0]$ & kg \\
    Box/board friction (static) & $[0.15,\,0.95]^3$ & -- \\
    Box/board friction (dynamic) & $[0.10,\,0.85]^3$ & -- \\
    Box/board restitution & $[0.0,\,0.05]^3$ & -- \\
    Planar offset (bottom, top) & $[-10,10]^2,\ [-5,5]^2$ & cm \\
    \hline
  \end{tabular}
\end{table}

\subsection{Observations and Privileged Information}

For TAR*, the actor observation is a 45-D proprioceptive vector consisting of base angular velocity, projected gravity, commanded base velocity, joint positions and velocities, and the previous action.
The recurrent encoder receives only the current observation at each step and maintains its history through the recurrent state, while the base linear velocity estimator additionally uses the latest four observations together with the 64-D recurrent latent $\hat{z}_t$ to estimate $\hat{v}_t$.
During training, the teacher encoder and AAC critic share the same 195-D privileged observation, containing ground-truth robot states, randomized dynamics parameters, foot contact forces, terrain heights, base linear acceleration, and the full states and physical parameters of both payloads.
The student encoder, in contrast, infers the latent solely from proprioception, and none of these privileged quantities are required at deployment.

PAMORT uses the same observation structure as TAR* and additionally conditions the actor on the 2-D preference $\boldsymbol{w}_t$; the recurrent encoder, teacher--student representation, and velocity estimator are unchanged.

\subsection{Network Architectures}

TAR* uses a one-layer LSTM student encoder with hidden size 256 to produce the 64-D latent $\hat{z}_t$, while its teacher encoder is an MLP with hidden dimensions $[256,128,64]$ that maps privileged observations to a 64-D latent.
A $[64,32]$ MLP estimates the 3-D base linear velocity.
The actor and critic are MLPs with hidden dimensions $[512,256,128]$ and ELU activations, and the actor outputs 12 joint-position actions.

PAMORT retains this architecture, adds the 2-D preference $\boldsymbol{w}_t$ to the actor input, and uses two critic outputs for the \texttt{move} and \texttt{payload} objectives.

WA uses actor and critic MLPs with hidden dimensions $[256,128]$ and $[512,256,128]$, respectively, with ELU activations.

\subsection{Training Hyperparameters}

TAR* and PAMORT use the same PPO settings: 24 steps per environment per iteration, five learning epochs with four mini-batches, Adam with a fixed learning rate of $1\times10^{-4}$, $\gamma=0.99$, GAE parameter $\lambda=0.95$, clipping parameter $0.2$, value-loss coefficient $1.0$, entropy coefficient $0.01$, and maximum gradient norm $1.0$.
The TAR auxiliary-loss coefficient is 1.0, and left--right symmetry augmentation is applied.

WA updates the preference every four low-level steps ($0.08\,\mathrm{s}$), treating them as one macro-transition.
It uses the same PPO settings except for a learning rate of $5\times10^{-4}$ and entropy coefficient of 0.005, with the macro-step discount factor derived consistently from the low-level $\gamma=0.99$.

\section{EXPERIMENTAL RESULTS}

\subsection{Multi-Objective Policy Analysis}
\label{sec:pareto}
We first examine whether PAMORT's MO base policy learns a diverse family of behaviors conditioned on the preference vector.
We compare it with TAR* and three fixed-weight TAR* variants, for which all reward weights assigned to the \texttt{payload} group in Table~\ref{tab:reward} are scaled by $0.75$, $0.5$, and $0.25$.

For this analysis, we use the six terrain families in Table~\ref{tab:eval0_terrains}, without separating the ascending and descending directions of the slope and stair terrains.
Each terrain is evaluated over all 10 difficulty levels, with the maximum difficulty set to $1.5$ times the training maximum.
For PAMORT, we evaluate 21 uniformly spaced preferences on the two-objective simplex,
with 50 randomized trials for each preference at each terrain and difficulty level.
The robot traverses a 10-m course with a commanded forward velocity of $1\,\mathrm{m/s}$.
A trial is considered successful if the robot reaches the goal with the payload retained, while payload drops and robot falls terminate the episode as failures.
For episodes longer than $10\,\mathrm{s}$, the accumulated rewards are rescaled to a 10\,s duration to avoid favoring slower policies.

All policies were trained for 130\,k iterations.
For both checkpoint selection and Pareto analysis, we use undiscounted episodic returns computed from unstandardized raw rewards and rescaled to a 10-s equivalent.
For TAR* and its fixed-weight variants, we selected the checkpoint maximizing the summed return over successful trials across all terrains and difficulty levels. For PAMORT, the \texttt{move} and \texttt{payload} returns, excluding the two box-drop rewards, were averaged over 50 trials for each terrain, level, and preference, and the empirical success rate was used to compute the ``expected'' hypervolume following~\cite{guerreiro2021hypervolume, yildiz2011union}; the checkpoint maximizing its sum over all terrains and levels was selected.
These checkpoints are used in all subsequent evaluations.

\begin{figure}[t]
    \centering
    \includegraphics[width=0.7\linewidth]{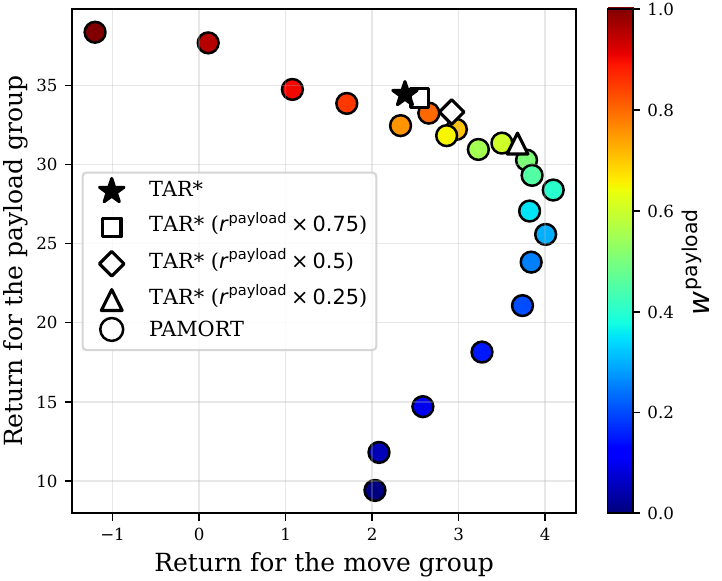}
    \caption{
        Empirical Pareto front of undiscounted episodic returns for the \texttt{move} and \texttt{payload} reward groups (upper right is better). PAMORT's MO policy yields non-dominated solutions for $0.4 \leq w^{\mathrm{payload}} \leq 1.0$.
    }
    \label{fig:eval3pf_pareto}
\end{figure}

Fig.~\ref{fig:eval3pf_pareto} shows these returns averaged over all trials, including failed trials, across the evaluated terrains and difficulty levels.
PAMORT's MO base policy exhibits a continuous trade-off between the two objectives as $w^{\mathrm{payload}}$ varies.
In particular, preferences in the range $0.4 \leq w^{\mathrm{payload}} \leq 1.0$ yield non-dominated behaviors, while the four independently trained TAR* policies lie close to a portion of this empirical Pareto front.
Thus, a single PAMORT training run recovers behaviors comparable to multiple TAR* policies trained with different fixed reward weights.

For $w^{\mathrm{payload}}<0.4$, the policies frequently drop the payload and terminate early, resulting in lower returns for both objectives.
Although such preferences are unsuitable as persistent operating points, the learned behavioral diversity provides the basis for the WA to select different move--payload trade-offs online.

\subsection{Simulation Performance}
\label{sec:sim}
Having established that the MO base policy represents a range of move--payload trade-offs, we next evaluate the task-level performance of the complete PAMORT framework, including the WA\@.
The WA is trained for 5,000 iterations, by which point its training has sufficiently converged.
We compare PAMORT with TAR* under identical robot, payload, and randomized physical conditions.

\begin{table}[t]
  \centering
  \caption{Terrain families and parameters for simulation evaluation}
  \label{tab:eval0_terrains}
  \begin{tabular}{ll}
    \hline
    Terrain Type & Parameter \\
    \hline
    RandomRough &
    $h_{\max}=0.15d_k\,\mathrm{m}$ \\
    CoarseBlocks &
    $l_{\mathrm{cell}}=0.9\,\mathrm{m},\;
     h_{\max}=0.15d_k\,\mathrm{m}$ \\
    FineBlocks &
    $l_{\mathrm{cell}}=0.45\,\mathrm{m},\;
     h_{\max}=0.15d_k\,\mathrm{m}$ \\
    SlopeUp/Down &
    $s=0.6d_k$ \\
    WideStairsUp/Down &
    $l_{\mathrm{tread}}=0.9\,\mathrm{m},\;
     h_{\mathrm{step}}=0.225d_k\,\mathrm{m}$ \\
    NarrowStairsUp/Down &
    $l_{\mathrm{tread}}=0.3\,\mathrm{m},\;
     h_{\mathrm{step}}=0.15d_k\,\mathrm{m}$ \\
    \hline
  \end{tabular}
\end{table}

\begin{figure}[t]
    \centering
    \includegraphics[width=\linewidth]{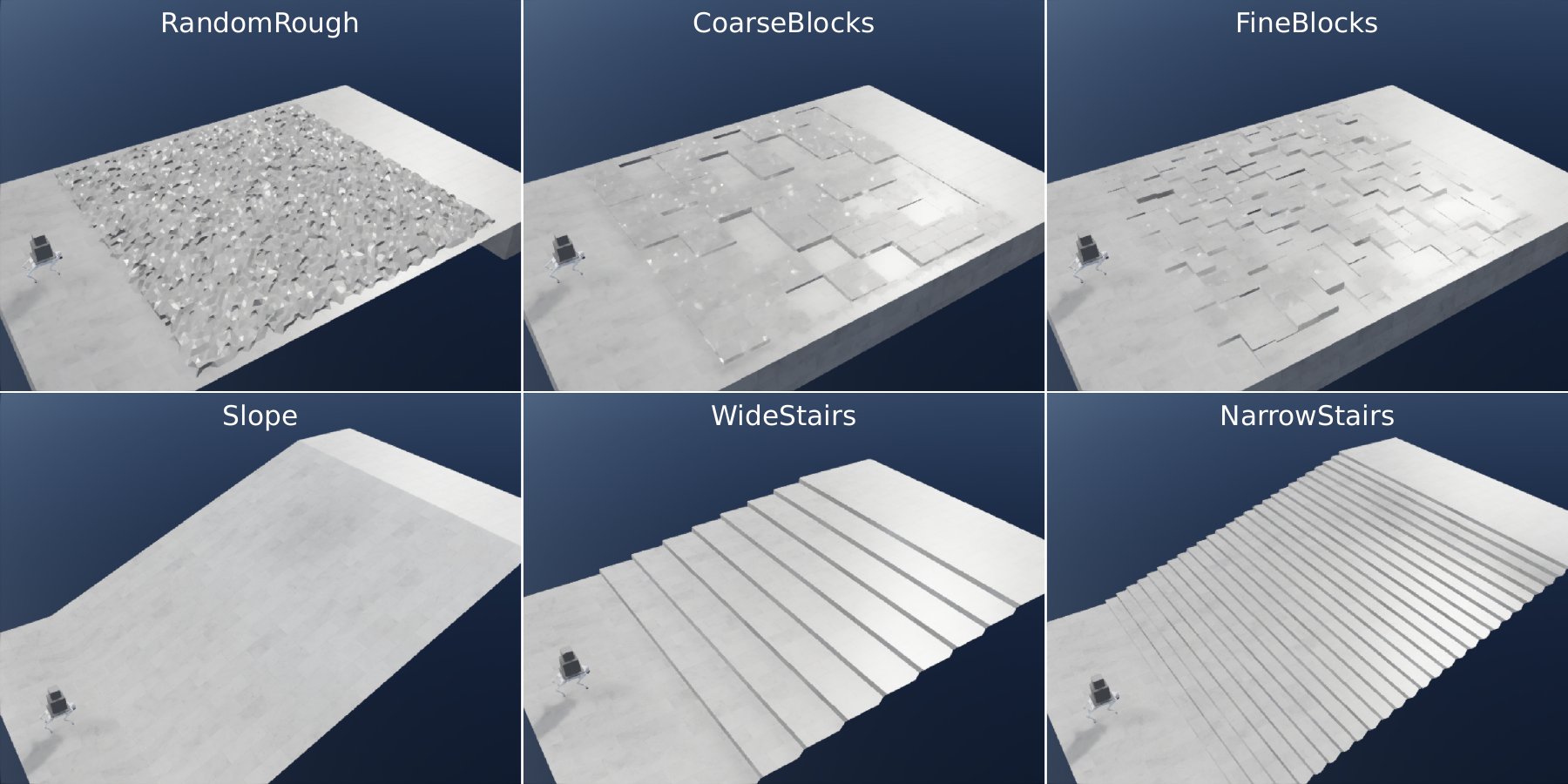}
    \caption{
        Representative level~6 simulation terrains. The ascending and descending variants share the same geometry and differ only in traversal direction.
    }
    \label{fig:eval0_terrains}
\end{figure}

We use the same six terrain families as in the preceding analysis and Table~\ref{tab:eval0_terrains}\@.
For task-level evaluation, however, the slope and two stair families are evaluated separately in the ascending and descending directions, yielding nine one-way terrain types in total.
These terrain families are the same as those used during training.
Fig.~\ref{fig:eval0_terrains} shows representative terrains at level~6.
Each trial commands a forward velocity of $1\,\mathrm{m/s}$ over $10\,\mathrm{m}$ and is counted as successful if the robot reaches the goal within $15\,\mathrm{s}$ without falling or dropping any carried payload.
Each terrain is evaluated independently at levels 0--9 with 100 trials per level.
Physical parameters are randomized within the training ranges, using the same 100 randomization patterns across all terrain types and levels.
Terrain difficulty follows $d_k=k/9$, with the maximum evaluation difficulty set to $1.5\times$ the training maximum.
Thus, level~6 corresponds to the maximum training difficulty, while levels~7--9 extend beyond the training range and include out-of-distribution (OOD) conditions.
We additionally evaluate zero-, one-, two-, and three-payload configurations. The two-payload case matches training; the zero- and one-payload cases remove payloads from this stack, while the three-payload case adds a third box on top with independently randomized size, mass, and friction drawn from the same ranges as the top box, yielding a stack configuration unseen during training.

\begin{figure}[t]
    \centering
    \includegraphics[width=\linewidth]{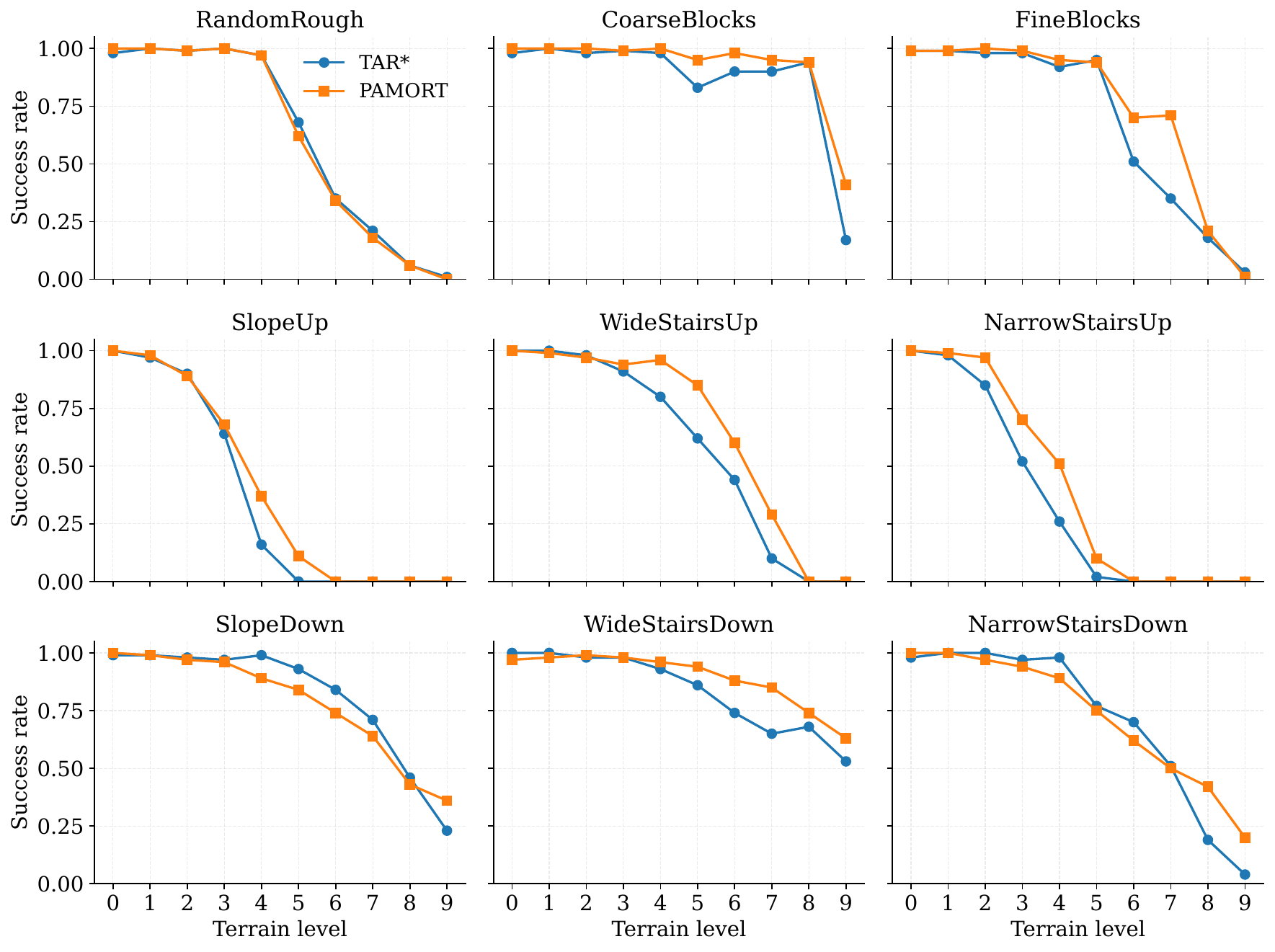}
    \caption{
      Success rates of TAR* and PAMORT across terrain difficulty levels under the two-payload training configuration. Each panel shows one of the nine one-way terrain types, with success rate over 100 trials at each level.
    }
\label{fig:eval0_success_rates}
\end{figure}

\begin{table*}[t]
  \centering
  \caption{
    Mean simulation success rates of TAR* (extension of TAR~\cite{mousa2025tar}) and PAMORT over levels 0--9 for different payload counts
  }
  \label{tab:payload_success_rates}
  \begin{tabular}{lcccccccc}
    \hline
    & \multicolumn{2}{c}{No payload}
    & \multicolumn{2}{c}{One payload}
    & \multicolumn{2}{c}{Two payloads (training setup)}
    & \multicolumn{2}{c}{Three payloads} \\
    \cline{2-9}
    Terrain
    & TAR* & PAMORT
    & TAR* & PAMORT
    & TAR* & PAMORT
    & TAR* & PAMORT \\
    \hline
    RandomRough
    & \textbf{0.699} & 0.670
    & \textbf{0.678} & 0.663
    & \textbf{0.625} & 0.616
    & \textbf{0.528} & 0.505 \\

    CoarseBlocks
    & 0.942 & \textbf{0.969}
    & 0.907 & \textbf{0.947}
    & 0.867 & \textbf{0.922}
    & 0.718 & \textbf{0.763} \\

    FineBlocks
    & 0.821 & \textbf{0.834}
    & 0.771 & \textbf{0.806}
    & 0.688 & \textbf{0.749}
    & 0.550 & \textbf{0.602} \\

    SlopeUp
    & 0.496 & \textbf{0.548}
    & 0.418 & \textbf{0.471}
    & 0.367 & \textbf{0.403}
    & 0.289 & \textbf{0.314} \\

    WideStairsUp
    & 0.695 & \textbf{0.737}
    & 0.648 & \textbf{0.694}
    & 0.585 & \textbf{0.660}
    & 0.461 & \textbf{0.492} \\

    NarrowStairsUp
    & 0.460 & \textbf{0.484}
    & 0.418 & \textbf{0.450}
    & 0.363 & \textbf{0.427}
    & 0.312 & \textbf{0.366} \\

    SlopeDown
    & 0.995 & \textbf{0.999}
    & 0.899 & \textbf{0.928}
    & \textbf{0.809} & 0.782
    & \textbf{0.698} & 0.668 \\

    WideStairsDown
    & \textbf{1.000} & \textbf{1.000}
    & 0.965 & \textbf{0.982}
    & 0.835 & \textbf{0.892}
    & 0.630 & \textbf{0.675} \\

    NarrowStairsDown
    & 0.930 & \textbf{0.966}
    & 0.799 & \textbf{0.880}
    & 0.714 & \textbf{0.729}
    & \textbf{0.586} & 0.567 \\
    \hline
    Mean
    & 0.782 & \textbf{0.801}
    & 0.723 & \textbf{0.758}
    & 0.650 & \textbf{0.687}
    & 0.530 & \textbf{0.550} \\
    \hline
  \end{tabular}
\end{table*}

Fig.~\ref{fig:eval0_success_rates} shows the success rates across terrain levels for all nine terrains under the two-payload training configuration.
PAMORT performs comparably to or better than TAR* except on SlopeDown and NarrowStairsDown, with particularly clear gains on difficult ascending terrains.
Table~\ref{tab:payload_success_rates} reports the mean success rate over levels 0--9 for zero, one, two, and three payloads.
Averaged over all nine terrains, PAMORT outperforms TAR* for every payload condition and retains a higher mean success rate as the payload count increases, indicating improved robustness to unseen payload configurations.

To further assess the quality of PAMORT's online preference selection, we compare it with fixed-preference references. A globally best fixed preference can achieve competitive aggregate performance, but identifying it requires an offline preference sweep over the evaluation conditions. An oracle that retrospectively selects the best fixed preference for each terrain and payload condition achieved 0.738 on average, while PAMORT reached 0.699, or 94.7\% of the oracle performance, without offline preference selection or condition-specific tuning. Notably, under the unseen three-payload condition, PAMORT achieved 0.550 versus 0.570 for the oracle, retaining 96.5\% of its performance.

\subsection{Real-World Performance}
\label{sec:real}
We evaluate zero-shot sim-to-real transfer on a Unitree Go2 over uneven terrain constructed from wooden pallets.
We compare PAMORT with TAR* under two payload configurations: the two stacked boxes used during training (Fig.~\ref{fig:task}-C) and an unseen configuration with three stacked boxes (Fig.~\ref{fig:task}-D).
Their physical parameters are listed in Table~\ref{tab:boxparam}\@.
For each payload configuration, we conduct ten trials on 38\% and 44\% slopes and $16\,\mathrm{cm}$ and $20\,\mathrm{cm}$ steps, with each trial including consecutive ascent and descent.
The 44\% slope and both steps exceed the maximum terrain difficulties used during training: a 40\% slope and a $15\,\mathrm{cm}$ step.

\begin{table}[t]
\caption{Physical parameters of real-world payloads}
\label{tab:boxparam}
\centering
\begin{tabular}{c l c l c}
\hline
Payload & Box & Mass [kg] & Static Friction & Size [cm] \\
\hline
Two & Top & 1.24 & 0.39 (vs. bottom) & $22 \times 22 \times 22$ \\
    & Bottom & 2.33 & 0.35 (vs. board) & $27 \times 27 \times 27$ \\
\hline
Three & Top & 1.25 & 0.41 (vs. mid) & $22 \times 22 \times 16$ \\
      & Mid & 1.43 & 0.40 (vs. bottom) & $34 \times 26 \times 18$ \\
      & Bottom & 0.86 & 0.36 (vs. board) & $28 \times 27 \times 20$ \\
\hline
\end{tabular}
\end{table}

As shown in Table~\ref{tab:expreal_success_rate}, PAMORT outperforms TAR* on three of the four terrain conditions with two payloads and achieves a higher mean success rate ($0.875$ vs.\ $0.650$). With three payloads, PAMORT again performs better on three terrains and achieves a higher mean success rate ($0.825$ vs.\ $0.700$). Across both payload configurations and all eight tasks, PAMORT achieves an overall mean success rate of $0.850$, compared with $0.675$ for TAR*. These results demonstrate robust zero-shot transfer to challenging terrains and a payload configuration not used during training.

\begin{table}[t]
  \centering
  \caption{
    Real-world success rates over 10 trials per task
  }
  \label{tab:expreal_success_rate}
  \begin{tabular}{lcccc}
    \hline
    & \multicolumn{2}{c}{Two payloads (training setup)}
    & \multicolumn{2}{c}{Three payloads} \\
    \cline{2-5}
    Terrain
    & TAR* & PAMORT
    & TAR* & PAMORT \\
    \hline
    38\% Slope
    & 0.6 & \textbf{1.0}
    & 0.7 & \textbf{0.9} \\
    44\% Slope
    & 0.5 & \textbf{0.9}
    & 0.6 & \textbf{0.9} \\
    $16\,\mathrm{cm}$ Step
    & 0.9 & \textbf{1.0}
    & 0.7 & \textbf{0.8} \\
    $20\,\mathrm{cm}$ Step
    & \textbf{0.6} & \textbf{0.6}
    & \textbf{0.8} & 0.7 \\
    \hline
    Mean
    & 0.650 & \textbf{0.875}
    & 0.700 & \textbf{0.825} \\
    \hline
  \end{tabular}
\end{table}

\begin{figure}[t]
    \centering
    \includegraphics[width=\linewidth]{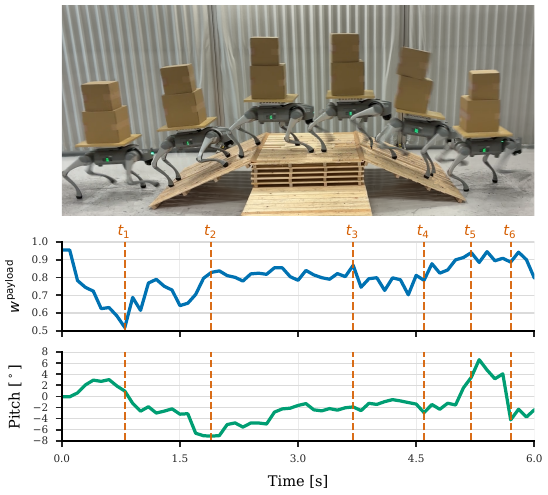}
    \caption{
    Payload preference $w^{\mathrm{payload}}$ and torso pitch during a representative two-payload traversal on a 44\% slope, which is slightly beyond the maximum training difficulty. Positive pitch denotes a nose-down posture, and $t_1$--$t_6$ correspond to the snapshots.
    }
    \label{fig:expreal_slope}
\end{figure}

To further examine how PAMORT adapts online, Fig.~\ref{fig:expreal_slope} shows the online preference adaptation of the WA during a representative traversal with two payloads on a 44\% ($24\tcdegree$) slope, slightly beyond the maximum training difficulty, at the maximum training command of $1.0\,\mathrm{m/s}$.
The payload preference $w^{\mathrm{payload}}$ is approximately $0.95$ while the robot is stationary at $t=0\,\mathrm{s}$. When the robot starts walking on flat ground, it decreases to approximately $0.5$--$0.7$, placing greater emphasis on velocity tracking (around $t_1$). During ascent ($t_2$--$t_3$), it increases to approximately $0.8$, and the robot reaches the top with little payload displacement. On the flat section at the top ($t_3$--$t_4$), it occasionally decreases to approximately $0.7$. It then increases to approximately $0.9$ during descent and landing ($t_4$--$t_6$). At the beginning of the descent ($t_4$--$t_5$), the robot accelerates and the payload temporarily shifts rearward, returning toward the center after landing ($t_6$). This rearward offset during $t_4$--$t_5$ may help compensate for the forward inertial motion induced by landing. 
Overall, the WA shifts the preference toward payload stabilization during the more demanding phases of traversal.

The torso pitch, defined as positive in the nose-down direction, remains within approximately $\pm7\tcdegree$ despite the $24\tcdegree$ slope. The robot pitches forward at startup ($t_1$) to suppress rearward payload motion and pitches backward after the descent ($t_6$) to suppress forward payload motion. These behaviors are consistent with the intended effect of the planar acceleration penalty in Eq.~\eqref{eq:planar_acc}.

On the real robot, policy inference takes $0.4\,\mathrm{ms}$ for TAR* and $0.5\,\mathrm{ms}$ for PAMORT, including the WA\@. Both are well below the $20\,\mathrm{ms}$ control period.

\section{CONCLUSIONS}
We proposed PAMORT for quadrupedal transportation of unsecured stacked payloads over uneven terrain. A preference-conditioned MO base policy represents locomotion--payload trade-offs, while a weight adjuster adapts the preference online from proprioception. In simulation, PAMORT outperforms TAR* on average across all evaluated payload configurations, including an unseen three-box stack. PAMORT also transfers zero-shot to a Unitree Go2, achieving a mean success rate of $0.850$ across eight real-world tasks compared with $0.675$ for TAR*. These results demonstrate robust payload transportation across terrain and payload conditions beyond those used during training.

The present study focuses on rigid, box-shaped payloads and considers only two objective groups, \emph{move} and \emph{payload}.
Important directions for future work include extending PAMORT to irregular, deformable, or liquid payloads and introducing additional objective groups to represent finer-grained trade-offs.
However, scaling to four or more objectives, a setting commonly referred to as many-objective optimization, makes learning diverse and well-converged Pareto-optimal behaviors increasingly difficult, particularly in MORL, where these behaviors must be acquired through interactions with the environment.
Developing scalable many-objective reinforcement learning methods therefore represents an important step toward more general robotic transportation.








\bibliographystyle{IEEEtran}
\bibliography{IEEEabrv,mybib}

\end{document}